\documentclass{article}

\usepackage{microtype}
\usepackage{graphicx}
\usepackage{subfigure}
\usepackage{booktabs} 
\usepackage{amsmath}
\usepackage{amsfonts}
\usepackage{nicefrac}
\usepackage{multirow}
\usepackage{multicol}
\usepackage{enumitem}
\usepackage{mathtools}
\usepackage[font=small,labelfont=bf,tableposition=top]{caption}

\DeclareCaptionLabelFormat{andtable}{#1~#2  \&  \tablename~\thetable}

\newcommand{\rmm}{\mathbf{m}}
\newcommand{\rmh}{\mathbf{h}}

\newcommand{\rmx}{\mathbf{x}}
\newcommand{\rmy}{\mathbf{y}}

\newcommand{\rmp}{\mathbf{p}}

\newcommand{\rmv}{\mathbf{v}}
\newcommand{\rmz}{\mathbf{z}}
\newcommand{\rmr}{\mathbf{r}}
\newcommand{\En}{\mathrm{E}(n)}
\newcommand{\summessages}{\rmm_i = \sum_{j \in \mathcal{N}(i)} \rmm_{ij}}
\newcommand{\summessagesb}{\rmm_i = \sum_{j \neq i} \rmm_{ij}}

\usepackage{hyperref}

\usepackage[accepted]{icml2021}

\icmltitlerunning{E(n) Equivariant Graph Neural Networks}

\begin{document}

\twocolumn[
\icmltitle{E(n) Equivariant Graph Neural Networks}



\icmlsetsymbol{equal}{*}

\begin{icmlauthorlist}
\icmlauthor{Victor Garcia Satorras}{delta}
\icmlauthor{Emiel Hoogeboom}{delta}
\icmlauthor{Max Welling}{delta}
\end{icmlauthorlist}

\icmlaffiliation{delta}{UvA-Bosch Delta Lab, University of Amsterdam, Netherlands}

\icmlcorrespondingauthor{Victor Garcia Satorras}{v.garciasatorras@uva.nl}
\icmlcorrespondingauthor{Emiel Hoogeboom}{e.hoogeboom@uva.nl}
\icmlcorrespondingauthor{Max Welling}{m.welling@uva.nl}

\icmlkeywords{Machine Learning, ICML}

\vskip 0.3in
]



\printAffiliationsAndNotice{}  

\begin{abstract}
This paper introduces a new model to learn graph neural networks equivariant to rotations, translations, reflections and permutations called $\En$-Equivariant Graph Neural Networks (EGNNs). In contrast with existing methods, our work does not require computationally expensive higher-order representations in intermediate layers while it still achieves competitive or better performance. In addition, whereas existing methods are limited to equivariance on $3$ dimensional spaces, our model is easily scaled to higher-dimensional spaces. We demonstrate the effectiveness of our method on dynamical systems modelling, representation learning in graph autoencoders and predicting molecular properties.
\end{abstract}

\begin{figure}
\begin{center}
\includegraphics[width=0.99\columnwidth]{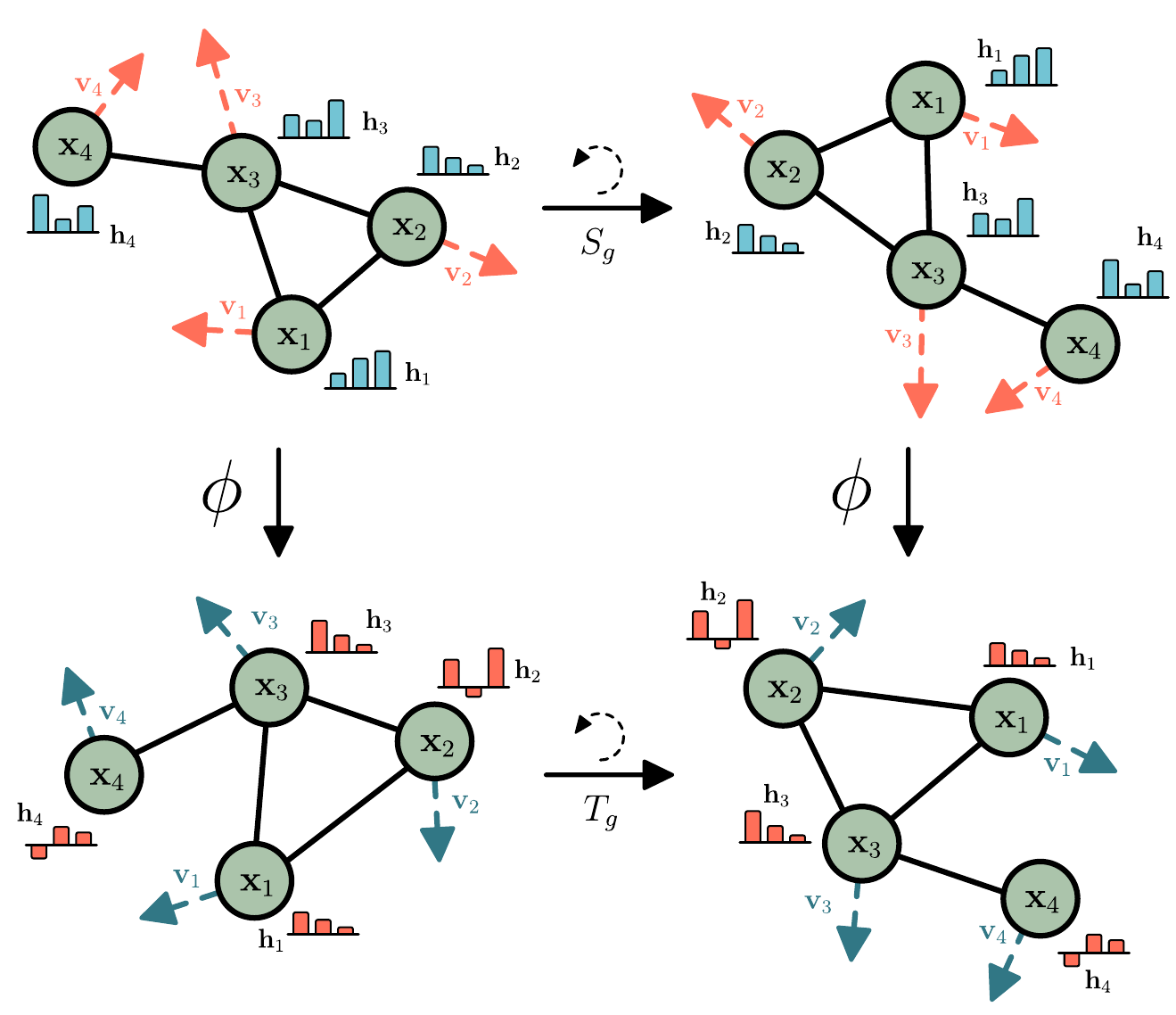}
\caption{Example of rotation equivariance on a graph with a graph neural network $\phi$}
\label{fig:overview_equivariance}
\end{center}
\end{figure}
\vspace{-5pt}

\section{Introduction}
Although deep learning has largely replaced hand-crafted features, many advances are critically dependent on inductive biases in deep neural networks. An effective method to restrict neural networks to relevant functions is to exploit the \textit{symmetry} of problems by enforcing equivariance with respect to transformations from a certain symmetry group. Notable examples are translation equivariance in Convolutional Neural Networks and permutation equivariance in Graph Neural Networks \cite{bruna2013spectral, defferrard2016convolutional, kipf2016semi}. 

Many problems exhibit 3D translation and rotation symmetries. Some examples are point clouds \cite{uy2019revisiting}, 3D molecular structures \cite{ramakrishnan2014quantum} or N-body particle simulations \cite{kipf2018neural}. The group corresponding to these symmetries is named the Euclidean group: SE(3) or when reflections are included E(3). It is often desired that predictions on these tasks are either equivariant or invariant with respect to E(3) transformations.

Recently, various forms and methods to achieve E(3) or SE(3) equivariance have been proposed \cite{thomas2018tensor, fuchs2020se, finzi2020generalizing, kohler2020equivariant_icml}. Many of these works achieve innovations in studying types of higher-order representations for intermediate network layers. However, the transformations for these higher-order representations require coefficients or approximations that can be expensive to compute. Additionally, in practice for many types of data the inputs and outputs are restricted to scalar values (for instance temperature or energy, referred to as type-$0$ in literature) and 3d vectors (for instance velocity or momentum, referred to as type-$1$ in literature). 



In this work we present a new architecture that is translation, rotation and reflection equivariant ($\En$), and permutation equivariant with respect to an input set of points. Our model is simpler than previous methods in that it does not require the spherical harmonics as in \cite{thomas2018tensor, fuchs2020se} while it can still achieve competitive or better results. In addition, equivariance in our model is not limited to the 3-dimensional space and can be scaled to larger dimensional spaces without a significant increase in computation. 

We evaluate our method in modelling dynamical systems, representation learning in graph autoencoders and predicting molecular properties in the QM9 dataset. Our method reports the best or very competitive performance in all three experiments.

\section{Background}
In this section we introduce the relevant materials on equivariance and graph neural networks which will later complement the definition of our method.

\subsection{Equivariance}
\label{sec:background_equivariance}
Let $T_g: X \xrightarrow{} X$ be a set of transformations on $X$ for the abstract group $g \in G$. We say a function $\phi: X \xrightarrow{} Y$ is equivariant to $g$ if there exists an equivalent transformation on its output space $S_g: Y \xrightarrow{} Y$  such that:
\begin{equation} \label{eq:equivariance}
    \phi(T_g(\rmx)) = S_g(\phi(\rmx)) 
\end{equation}
As a practical example, let $\phi(\cdot)$ be a non-linear function, $\rmx = (\rmx_1, \dots, \rmx_M) \in \mathbb{R}^{M \times n}$ an input set of $M$ point clouds embedded in a $n$-dimensional space, $\phi(\rmx) = \rmy \in \mathbb{R}^{M \times n}$ the transformed set of point clouds, $T_g$ a translation on the input set $T_g(\rmx) = \rmx + g$ and $S_g$ an equivalent translation on the output set $S_g(\rmy) = \rmy + g$. If our transformation $\phi: X \xrightarrow{} Y$ is translation equivariant, translating the input set $T_g(\rmx)$ and then applying the function $\phi(T_x(\rmx))$ on it, will deliver the same result as first running the function $y=\phi(\rmx)$ and then applying an equivalent translation to the output $T_g(\rmy)$ such that Equation \ref{eq:equivariance} is fulfilled and $\phi(\rmx + g) = \phi(\rmx) + g$. In this work we explore the following three types of equivariance on a set of particles $\rmx$:
\begin{enumerate}[nolistsep]
    \item Translation equivariance. Translating the input by $g \in \mathbb{R}^n$ results in an equivalent translation of the output. Let $\rmx + g$ be shorthand for $(\rmx_1 + g, \ldots, \rmx_M + g)$. Then $\rmy + g = \phi(\rmx + g)$ 
    \item Rotation (and reflection) equivariance. For any orthogonal matrix $Q \in \mathbb{R}^{n \times n}$, let $Q\rmx$ be shorthand for $(Q\rmx_1, \ldots, Q\rmx_M)$. Then rotating the input results in an equivalent rotation of the output $Q\rmy = \phi(Q\rmx)$.
    \item Permutation equivariance. Permuting the input results in the same permutation of the output $P(\rmy) = \phi(P(\rmx))$ where $P$ is a permutation on the row indexes.
\end{enumerate}

Note that velocities $\rmv 
\in \mathbb{R}^{M \times n}$ are unaffected by translations, but they transform equivalently under rotation (2) and permutation~(3). Our method introduced in Section \ref{sec:method_main} will satisfy the three above mentioned equivariant constraints. 

\subsection{Graph Neural Networks}
\label{sec:background_gnn}
Graph Neural Networks are permutation equivariant networks that operate on graph structured data \cite{bruna2013spectral, defferrard2016convolutional, kipf2016semi}. Given a graph $\mathcal{G} = (\mathcal{V}, \mathcal{E})$ with nodes $v_i \in \mathcal{V}$ and edges $e_{ij} \in \mathcal{E}$ we define a graph convolutional layer following notation from \cite{gilmer2017neural} as:

\begin{equation} \label{eq:gnn}
\begin{aligned}
\rmm_{ij} &= \phi_e(\rmh_i^l, \rmh_j^l, a_{ij}) \\
    \rmm_i &= \sum_{j \in \mathcal{N}(i)} \rmm_{i j} \\
    \rmh_i^{l+1} &= \phi_h(\rmh_i^l, \rmm_i) \\
\end{aligned}
\end{equation}

Where $\rmh_i^l \in \mathbb{R}^{\text{nf}}$ is the nf-dimensional embedding of node $v_i$ at layer $l$. $a_{ij}$ are the edge attributes. $\mathcal{N}(i)$ represents the set of neighbors of node $v_i$. Finally, $\phi_e$ and $\phi_h$ are the edge and node operations respectively which are commonly approximated by Multilayer Perceptrons (MLPs).
\section{Equivariant Graph Neural Networks} 

\label{sec:method_main}
In this section we present Equivariant Graph Neural Networks (EGNNs). Following the notation from background Section \ref{sec:background_gnn}, we consider a graph $\mathcal{G} = (\mathcal{V},\mathcal{E})$ with nodes $v_i \in \mathcal{V}$ and edges $e_{ij} \in \mathcal{E}$. In addition to the feature node embeddings $\rmh_i \in \mathbb{R}^{\text{nf}}$ we now also consider a $n$-dimensional coordinate $\rmx_i \in  \mathbb{R}^{\text{n}}$ associated with each of the graph nodes. Our model will preserve equivariance to rotations and translations on these set of coordinates $\rmx_i$ and it will also preserve equivariance to permutations on the set of nodes $\mathcal{V}$ in the same fashion as GNNs.



Our Equivariant Graph Convolutional Layer (EGCL) takes as input the set of node embeddings $\rmh^l =\{\rmh_0^l, \dots, \rmh_{M-1}^l\}$, coordinate embeddings $\rmx^l =\{\rmx_0^l, \dots, \rmx_{M-1}^l\} $ and edge information $\mathcal{E}=(e_{ij})$ and outputs a transformation on $\rmh^{l+1}$ and $\rmx^{l+1}$. Concisely: $\rmh^{l+1}, \rmx^{l+1} = \text{EGCL}[\rmh^{l}, \rmx^{l}, \mathcal{E}]$. The equations that define this layer are the following:
\begin{align}\label{eq:method_edge}
\rmm_{ij} &=\phi_{e}\left(\rmh_{i}^{l}, \rmh_{j}^{l},\left\|\rmx_{i}^{l}-\rmx_{j}^{l}\right\|^{2}, a_{i j}\right) \\ \label{eq:method_coords}
\rmx_{i}^{l+1} &=\rmx_{i}^{l}+ C\sum_{j \neq i}\left(\rmx_{i}^{l}-\rmx_{j}^{l}\right) \phi_{x}\left(\rmm_{ij}\right) \\
 \label{eq:method_agg}
\rmm_{i} &=\sum_{j \neq i} \rmm_{ij} \\ 
\label{eq:method_node}
\rmh_{i}^{l+1} &=\phi_{h}\left(\rmh_{i}^l, \rmm_{i}\right)
\end{align}


 
Notice the main differences between the above proposed method and the original Graph Neural Network from equation \ref{eq:gnn} are found in equations \ref{eq:method_edge} and \ref{eq:method_coords}. In equation \ref{eq:method_edge} we now input the relative squared distance between two coordinates $\|\mathbf{x}_{i}^{l}-\mathbf{x}_{j}^{l}\|^{2}$ into the edge operation $\phi_e$. The embeddings $\rmh_i^l$, $\rmh_j^l$, and the edge attributes $a_{ij}$ are also provided as input to the edge operation as in the GNN case. In our case the edge attributes will incorporate the edge values $a_{ij} = e_{ij}$, but they can also include additional edge information.

In Equation \ref{eq:method_coords} we update the position of each particle $\rmx_i$ as a vector field in a radial direction. In other words, the position of each particle $\rmx_i $ is updated by the weighted sum of all relative differences $(\rmx_i - \rmx_j)_{\forall j}$. The weights of this sum are provided as the output of the function $\phi_x: \mathbb{R}^{\text{nf}} \rightarrow \mathbb{R}^1$ that takes as input the edge embedding $\rmm_{ij}$ from the previous edge operation and outputs a scalar value. $C$ is chosen to be $1/(M-1)$, which divides the sum by its number of elements. This equation is the main difference of our model compared to standard GNNs and it is the reason why equivariances 1, 2 are preserved (proof in Appendix \ref{sec:appendix_equiv_proof}). Despite its simplicity, this equivariant operation is very flexible since the embedding $\rmm_{ij}$ can carry information from the whole graph and not only from the given edge $e_{ij}$. 

Finally, equations \ref{eq:method_agg} and \ref{eq:method_node} follow the same updates than standard GNNs. Equation \ref{eq:method_agg} is the aggregation step, in this work we choose to aggregate messages from all other nodes $j\neq i$, but we could limit the message exchange to a given neighborhood $j \in \mathcal{N}(i)$ if desired in both equations \ref{eq:method_agg} and \ref{eq:method_coords}. Equation \ref{eq:method_node} performs the node operation $\phi_h$ which takes as input the aggregated messages $\rmm_i$, the node emedding $\rmh_i^{l}$ and outputs the updated node embedding $\rmh_i^{l+1}$.


\subsection{Analysis on E(n) equivariance} \label{sec:equivariance_proof}
In this section we analyze the equivariance properties of our model for $E(3)$ symmetries (i.e. properties 1 and 2 stated in section \ref{sec:background_equivariance}). In other words, our model should be translation equivariant on $\rmx$ for any translation vector $g \in \mathbb{R}^n$ and it should also be rotation and reflection equivariant on $\rmx$ for any orthogonal matrix $Q \in \mathbb{R}^{n \times n}$. More formally our model satisfies:
\begin{equation*}
    Q\rmx^{l+1} + g, \rmh^{l+1}  = \mathrm{EGCL}(Q\rmx^l + g, \rmh^l)
\end{equation*}
We provide a formal proof of this in Appendix \ref{sec:appendix_equiv_proof}. Intuitively, let's consider a $\rmh^l$ feature which is already $\En$ invariant, then we can see that the resultant edge embedding $\rmm_{ij}$ from Equation \ref{eq:method_edge} will also be $\En$ invariant, because in addition to $\rmh^l$, it only depends on squared distances $\|\rmx_{i}^l-\rmx_{j}^l\|^{2}$, which are $\En$ invariant. Next, Equation \ref{eq:method_coords} computes $\rmx^{l+1}_i$ by a weighted sum of differences $(\rmx_i - \rmx_j)$ which is added to $\rmx_i$, this transforms as a type-1 vector and preserves equivariance (see Appendix \ref{sec:appendix_equiv_proof}). Finally the last two equations \ref{eq:method_agg} and \ref{eq:method_node} that generate the next layer node-embeddings $\rmh^{l+1}$ remain $\En$ invariant since they only depend on $\rmh^l$ and $\rmm_{ij}$ which, as we saw above, are $\En$ invariant. Therefore the output $\rmh^{l+1}$ is $\En$ invariant and $\rmx^{l+1}$ is $\En$ equivariant to $\rmx^{l}$. Inductively, a composition of $\mathrm{EGCL}$s will also be equivariant.

\subsection{Extending EGNNs for vector type representations} \label{sec:method_velocity}
In this section we propose a slight modification to the presented method such that we explicitly keep track of the particle's momentum. In some scenarios this can be useful not only to obtain an estimate of the particle's velocity at every layer but also to provide an initial velocity value in those cases where it is not 0. We can include momentum to our proposed method by just replacing Equation \ref{eq:method_coords} of our model with the following equation:
\begin{equation} \label{eq:egnn_velocity}
    \begin{aligned}
    \mathbf{v}_{i}^{l+1}&= \phi_{v}\left(\rmh_{i}^l\right)\rmv_{i}^\text{init} + C\sum_{j \neq i}\left(\mathbf{x}_{i}^{l}-\mathbf{x}_{j}^{l}\right) \phi_{x}\left(\mathbf{m}_{ij}\right) \\
 \rmx_{i}^{l+1} &=\mathbf{x}_{i}^{l}+ \mathbf{v}_i^{l+1}
    \end{aligned}
\end{equation}
Note that this extends the EGCL layer as $\rmh^{l+1}, \rmx^{l+1}, \rmv^{l+1} = \text{EGCL}[\rmh^{l}, \rmx^{l}, \rmv^{\text{init}}, \mathcal{E}]$. The only difference is that now we broke down the coordinate update (eq. \ref{eq:method_coords}) in two steps, first we compute the velocity $\rmv^{l+1}_i$ and then we use this velocity to update the position $\rmx^l_i$. The initial velocity $\rmv^{\text{init}}_i$ is scaled by a new function $\phi_v: \mathbb{R}^N \rightarrow \mathbb{R}^1$ that maps the node embedding $\rmh_i^l$ to a scalar value. Notice that if the initial velocity is set to zero (${\rmv}^{\text{init}}_i = 0$), both equations \ref{eq:method_coords} and \ref{eq:egnn_velocity} become exactly the same. We proof the equivariance of this variant of the model in Appendix \ref{sec:appendix_equiv_proof_velocity}. This variant is used in experiment \ref{sec:experiment_n_body} where we provide the initial velocity of the system, and predict a relative position change.

\begin{table*}[t] \tiny
  \centering
  \begin{tabular}{|l | c | c | c | c | c | c}
  \toprule
     & GNN & Radial Field & TFN & Schnet  & EGNN \\
    \midrule
    \multirow{2}*{Edge} & 
    \multirow{2}*{$\rmm_{ij} = \phi_e(\rmh_i^l, \rmh_j^l, a_{ij})$} &
    \multirow{2}*{$\rmm_{ij} = \phi_{\text{rf}}(\|\rmr_{ij}^{l}\|) \rmr_{ij}^{l}$} &
    \multirow{2}*{$\rmm_{ij} =  \sum_{k}\mathbf{W}^{lk}\rmr_{ji}^{l}\rmh_i^{lk}$} &
    \multirow{2}*{$\rmm_{ij}=\phi_{\text{cf}}(\|\rmr_{ij}^{l}\|) \phi_\text{s}(\rmh_{j}^{l})$ } & 
    $\mathbf{m}_{ij} =\phi_{e}(\mathbf{h}_{i}^{l}, \mathbf{h}_{j}^{l},\|\rmr_{ij}^{l}\|^{2}, a_{i j})$   \\
    & &  & & & $\hat{\rmm}_{ij} = \rmr_{ij}^{l}\phi_x(\rmm_{ij})$  \\
    \midrule
    \multirow{2}*{Agg}. &
    \multirow{2}*{$\summessages$} &
    \multirow{2}*{$\summessagesb$} &
    \multirow{2}*{$\summessagesb$} &
    \multirow{2}*{$\summessagesb$} &
    $\summessagesb$ \\
     & & & & &
     $\hat{\rmm}_i = C\sum_{j \neq i} \hat{\rmm}_{ij}$ \\
    \midrule
    \multirow{2}*{Node} &
    \multirow{2}*{$\rmh_i^{l+1} = \phi_h(\rmh_i^l, \rmm_i)$} &
    \multirow{2}*{$\rmx_i^{l+1} = \rmx_i^{l} + \rmm_i$} &
    \multirow{2}*{$\rmh_i^{l+1} = w^{ll}\rmh_i^{l} + \rmm_i$} &
    \multirow{2}*{$\rmh_i^{l+1} = \phi_h(\rmh_i^l, \rmm_i)$} &
    $\mathbf{h}_{i}^{l+1} =\phi_{h}\left(\mathbf{h}_{i}^l, \mathbf{m}_{i}\right)$ \\
         &
     &
     &
     &
     &
    $\rmx_i^{l+1} = \rmx_i^{l} + \hat{\rmm}_i $  \\
    \midrule
     &
    Non-equivariant &
    $\En$-Equivariant &
    $\mathrm{SE}(3)$-Equivariant &
    $\En$-Invariant &
    $\En$-Equivariant  \\
    \bottomrule
  \end{tabular}
  \caption{Comparison over different works from the literature under the message passing framework notation. We created this table with the aim to provide a clear and simple way to compare over these different methods. The names from left to right are: Graph Neural Networks \cite{gilmer2017neural}; Radial Field from Equivariant Flows \cite{kohler2019equivariant_workshop}; Tensor Field Networks \cite{thomas2018tensor}; Schnet \cite{schutt2017schnet}; and our Equivariant Graph Neural Network. The difference between two points is written $\rmr_{ij} = (\rmx_i - \rmx_j)$.}
  \label{tab:related_work}
\end{table*}

\subsection{Inferring the edges} \label{sec:edges_inference}
Given a point cloud or a set of nodes, we may not always be provided with an adjacency matrix. In those cases we can assume a fully connected graph where all nodes exchange messages with each other $j \neq i$ as done in Equation \ref{eq:method_agg}. This fully connected approach may not scale well to large point clouds where we may want to locally limit the exchange  of messages $\rmm_i = 
   \sum_{j \in \mathcal{N}(i)}\rmm_{ij}$ to a neighborhood $\mathcal{N}(i)$ to avoid an overflow of information.

Similarly to \cite{serviansky2020set2graph, kipf2018neural}, we present a simple solution to infer the relations/edges of the graph in our model, even when they are not explicitly provided. Given a set of neighbors $\mathcal{N}(i)$ for each node $i$, we can re-write the aggregation operation from our model (eq. \ref{eq:method_agg}) in the following way:
\begin{equation}\label{eq:attention}
   \rmm_i = 
   \sum_{j \in \mathcal{N}(i)}\rmm_{ij} = \sum_{j \neq i} e_{ij}\rmm_{ij} 
\end{equation}
Where $e_{ij}$ takes value $1$ if there is an edge between nodes $(i,j)$ and $0$ otherwise. 
Now we can choose to approximate the relations $e_{ij}$ with the following function $e_{ij} \approx \phi_{inf}(\rmm_{ij})$, where $\phi_{inf}: \mathbb{R}^{nf} \rightarrow [0, 1]^1$ resembles a linear layer followed by a sigmoid function that takes as input the current edge embedding and outputs a soft estimation of its edge value. This modification does not change the $\En$ properties of the model since we are only operating on the messages $\rmm_{ij}$ which are already $\En$ invariant. 

\section{Related Work} \label{sec:related_work}
Group equivariant neural networks have demonstrated their effectiveness in a wide variety of tasks \citep{cohen2016group,cohen2017steerable,weiler2019e2equivariant,rezende2019equivariantflows,romero2021group}.
Recently, various forms and methods to achieve E(3) or SE(3) equivariance have been proposed. \citet{thomas2018tensor, fuchs2020se} utilize the spherical harmonics to compute a basis for the transformations, which allows transformations between higher-order representations. A downside to this method is that the spherical harmonics need to be recomputed which can be expensive. Currently, an extension of this method to arbitrary dimensions is unknown. \citet{finzi2020generalizing} parametrize transformations by mapping kernels on the Lie Algebra. For this method the neural network outputs are in certain situations stochastic, which may be undesirable. \citet{horie2020isometric} proposes a set of isometric invariant and equivairant transformations for Graph Neural Networks.
\citet{kohler2019equivariant_workshop, kohler2020equivariant_icml} propose an $\En$ equivariant network to model 3D point clouds, but the method is only defined for positional data on the nodes without any feature dimensions.

Another related line of research concerns message passing algorithms on molecular data. \cite{gilmer2017neural} presented a message passing setting (or Graph Neural Network) for quantum chemistry, this method is permutation equivariant but not translation or rotation equivariant. \cite{kondor2018covariant} extends the equivariance of GNNs such that each neuron transforms in a specific way under permutations, but this extension only affects its permutation group and not translations or rotations in a geometric space. Further works \cite{schutt2017schnet, schutt2017quantum} build $\En$ invariant message passing networks by inputting the relative distances between points. \citet{klicpera2020directional, klicpera2020fast} in addition to relative distances it includes a modified message passing scheme analogous to Belief Propagation that considers angles and directional information equivariant to rotations. It also uses Bessel functions and spherical harmonics to construct and orthogonal basis. \citet{anderson2019cormorant, miller2020relevance} include $SO(3)$ equivariance in its intermediate layers for modelling the behavior and properties of molecular data. Our method is also framed as a message passing framework but in contrast to these methods it achieves $\En$ equivariance.

\subsection*{Relationship with existing methods}
 In Table \ref{tab:related_work} the EGNN equations are detailed together with some of its closest methods from the literature under the message passing notation from \cite{gilmer2017neural}. This table aims to provide a simple way to compare these different algorithms. It is structured in three main rows that describe i) the edge ii) aggregation and iii) node update operations. The GNN algorithm is the same as the previously introduced in Section \ref{sec:background_equivariance}. Our EGNN algorithm is also equivalent to the description in Section \ref{sec:method_main} but notation has been modified to match the (edge, aggregation, node) format. In all equations $\rmr_{ij}^l = (\rmx_i - \rmx_j)^l$. Notice that except the EGNN, all algorithms have the same aggregation operation and the main differences arise from the edge operation. The algorithm that we call "Radial Field" is the $\En$ equivariant update from \cite{kohler2019equivariant_workshop}. This method is $\En$ equivariant, however its main limitation is that it only operates on $\rmx$ and it doesn't propagate node features $\rmh$ among nodes. In the method $\phi_{rf}$ is modelled as an MLP. Tensor Field Networks (TFN) \cite{thomas2018tensor} instead propagate the node embeddings $\rmh$ but it uses spherical harmonics to compute its learnable weight kernel $ \mathbf{W}^{\ell k}: \mathbb{R}^{3} \rightarrow \mathbb{R}^{(2 \ell+1) \times(2 k+1)}$ which preserves $SE(3)$ equivariance but is expensive to compute an limited to the 3 dimensional space. The SE(3) Transformer \cite{fuchs2020se} (not included in this table), can be interpreted as an extension of TFN with attention. Schnet \cite{schutt2017schnet} can be interpreted as an $\En$ invariant Graph Neural Network where $\phi_{cf}$ receives as input relative distances and outputs a continuous filter convolution that multiplies the neighbor embeddings $\rmh$. Our EGNN differs from these other methods in terms that it performs two different updates in each of the table rows, one related to the embeddings $\rmh$ and another related to the coordinates $\rmx$, these two variables exchange information in the edge operation. In summary the EGNN can retain the flexibility of GNNs while remaining $\En$ equivariant as the Radial Field algorithm and without the need to compute expensive operations (i.e. spherical harmonics).

\section{Experiments}

\subsection{Modelling a dynamical system | N-body system } \label{sec:experiment_n_body}

In a dynamical system a function defines the time dependence of a point or set of points in a geometrical space. Modelling these complex dynamics is crucial in a variety of applications such as control systems \cite{chua2018deep}, model based dynamics in reinforcement learning \cite{nagabandi2018neural}, and physical systems simulations \cite{grzeszczuk1998neuroanimator, watters2017visual}. In this experiment we forecast the positions for a set of particles which are modelled by simple interaction rules, yet can exhibit complex dynamics.

Similarly to \cite{fuchs2020se}, we extended the Charged Particles N-body experiment from \cite{kipf2018neural} to a 3 dimensional space. The system consists of 5 particles that carry a positive or negative charge and have a position and a velocity associated in 3-dimensional space. The system is controlled by physic rules: particles are attracted or repelled depending on their charges. This is an equivariant task since rotations and translations on the input set of particles result in the same transformations throughout the entire trajectory.

\textbf{Dataset}: We sampled 3.000 trajectories for training, 2.000 for validation and 2.000 for testing. Each trajectory has a duration of 1.000 timesteps. For each trajectory we are provided with the initial particle positions $\rmp^{(0)}=\{ \rmp^{(0)}_1, \dots \rmp^{(0)}_5 \} \in \mathbb{R}^{5 \times 3}$, their initial velocities $\rmv^{(0)}=\{ \rmv^{(0)}_1, \dots \rmv^{(0)}_5 \} \in \mathbb{R}^{5 \times 3}$ and their respective charges $\mathbf{c} = \{c_1, \dots c_5\} \in \{-1, 1\}^5$. The task is to estimate the positions of the five particles after 1.000 timesteps. We optimize the averaged Mean Squared Error of the estimated position with the ground truth one.

\textbf{Implementation details:} In this experiment we used the extension of our model that includes velocity from section \ref{sec:method_velocity}. We input the position $\rmp^{(0)}$ as the first layer coordinates $\rmx^0$ of our model and the velocity $\rmv^{(0)}$ as the initial velocity in Equation \ref{eq:egnn_velocity}, the norms $\|\rmv_i^{(0)} \|$ are also provided as features to $\rmh_i^0$ through a linear mapping. The charges are input as edge attributes $a_{ij} = c_i c_j$. The model outputs the last layer coordinates $\rmx^L$ as the estimated positions.
We compare our method to its non equivariant Graph Neural Network (GNN) cousin, and the equivariant methods: Radial Field \cite{kohler2019equivariant_workshop}, Tensor Field Networks and the SE(3) Transformer. All algorithms are composed of 4 layers and have been trained under the same conditions, batch size 100, 10.000 epochs, Adam optimizer, the learning rate was tuned independently for each model. We used 64 features for the hidden layers in the Radial Field, the GNN and our EGNN. As non-linearity we used the Swish activation function \cite{ramachandran2017searching}. For TFN and the SE(3) Transformer we swept over different number of vector types and features and chose those that provided the best performance. Further implementation details are provided in Appendix \ref{sec:appendix_implementation_dynamical}. A Linear model that simply considers the motion equation $\rmp^{(t)} = \rmp^{(0)} + \rmv^{(0)}t$ is also included as a baseline. We also provide the average forward pass time in seconds for each of the models for a batch of 100 samples in a GTX 1080 Ti GPU.


\begin{table}[h] 
  \centering
  \begin{tabular}{lcc}
  \toprule
    Method & MSE & Forward time (s)  \\
    \midrule
    Linear & 0.0819 & .0001  \\
       SE(3) Transformer & 0.0244 & .1346  \\ 
       Tensor Field Network & 0.0155 & .0343  \\ 
        Graph Neural Network &  0.0107 & .0032  \\ 
       Radial Field & 0.0104 & .0039  \\
    \textbf{EGNN} & \textbf{0.0071}  & .0062\\ 
    \bottomrule
  \end{tabular}
  \caption{Mean Squared Error for the future position estimation in the N-body system experiment, and forward time in seconds for a batch size of 100 samples running in a GTX 1080Ti GPU.}
  \label{table:n_body}
\end{table}
\vspace{-5pt}

\textbf{Results}
As shown in Table \ref{table:n_body} our model significantly outperforms the other equivariant and non-equivariant alternatives while still being efficient in terms of running time. It reduces the error with respect to the second best performing method by a $32\%$. In addition it doesn't require the computation of spherical harmonics which makes it more time efficient than Tensor Field Networks and the SE(3) Transformer.

\begin{figure}[h!]
\begin{center}
\centerline{\includegraphics[width=0.8\columnwidth]{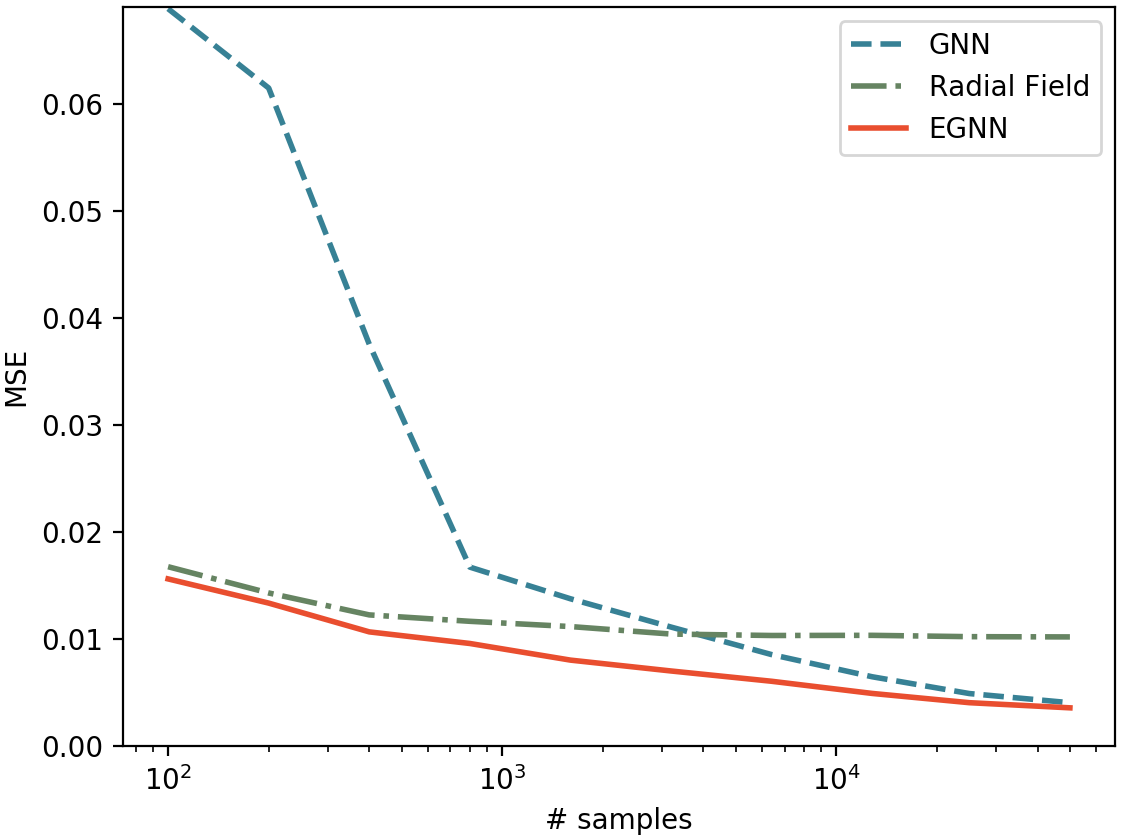}}
\caption{Mean Squared Error in the N-body experiment for the Radial Field, GNN and EGNN methods when sweeping over different amounts of training data.}
\label{fig:n_body_figure}
\end{center}
\vskip -0.25in
\end{figure}

\textbf{Analysis for different number of training samples:} We want to analyze the performance of our EGNN in the small and large data regime. In the following, we report on a similar experiment as above, but instead of using 3.000 training samples we generated a new training partition of 50.000 samples and we sweep over different amounts of data from 100 to 50.000 samples. We compare the performances of our EGNN vs its non-equivariant GNN counterpart and the Radial Field algorithm. Results are presented in Figure \ref{fig:n_body_figure}. Our method outperforms both Radial Field and GNNs in the small and large data regimes. This shows the EGNN is more data efficient than GNNs since it doesn't require to generalize over rotations and translations of the data while it ensembles the flexibility of GNNs in the larger data regime. Due to its high model bias, the Radial Field algorithm performs well when data is scarce but it is unable to learn the subtleties of the dataset as we increase the training size. In summary, our EGNN benefits from both the high bias of $\En$ methods and the flexibility of GNNs.

\subsection{Graph Autoencoder} \label{sec:expriment_autoencoder}

A Graph Autoencoder can learn unsupervised representations of graphs in a continuous latent space \cite{kipf2016variational, simonovsky2018graphvae}. In this experiment section we use our EGNN to build an Equivariant Graph Autoencoder. We will explain how Graph Autoencoders can benefit from equivariance and we will show how our method outperforms standard GNN autoencoders in the provided datasets. This problem is particularly interesting since the embedding space can be scaled to larger dimensions and is not limited to a 3 dimensional Euclidean space.

Similarly to the work of \cite{kipf2016variational} further extended by section 3.3 in \cite{liu2019graph}, our graph auto-encoder $\rmz = q(\mathcal{G})$ embeds a graph $\mathcal{G}$ into a set of latent nodes $\rmz = \{\rmz_1, \dots \rmz_M \} \in \mathbb{R}^{M \times n}$, where $M$ is the number of nodes and $n$ the embedding size per node. Notice this may reduce the memory complexity to store the graphs from $O(M^2)$ to $O(Mn)$ where $n$ may depend on $M$ for a certain approximation error tolerance. This differs from the variational autoencoder proposed in \cite{simonovsky2018graphvae} which embeds the graph in a single vector $\rmz \in \mathbb{R}^K$, which causes the reconstruction to be computationally very expensive since the nodes of the decoded graph have to be matched again to the ground truth. In addition to the introduced graph generation and representation learning methods, it is worth it mentioning that in the context of graph compression other methods \cite{candes2009exact} can be used.

More specifically, we will compare our Equivariant Graph Auto-Encoder in the task presented in \cite{liu2019graph} where a graph $\mathcal{G} = (\mathcal{V}, \mathcal{E})$ with node features $H \in \mathbb{R}^{M \times \text{nf}}$ and adjacency matrix $A \in \{0, 1\}^{M \times M}$ is embedded into a latent space $\rmz = q(H, A) \in \mathbb{R}^{M \times n}$. Following \cite{kipf2016variational, liu2019graph}, we are only interested in reconstructing the adjacency matrix $A$ since the datasets we will work with do not contain node features. The decoder $g(\cdot)$ proposed by \cite{liu2019graph} takes as input the embedding space $\rmz$ and outputs the reconstructed adjacency matrix $\hat{A} = g(\rmz)$, this decoder function is defined as follows:
\begin{equation} \label{eq:graph_decoder}
    \hat{A}_{ij} = g_e(\rmz_i, \rmz_j)=\frac{1}{1 + \mathrm{exp}(
w\left\|\rmz_{i}- \rmz_{j}\right\|^{2} + b
)}
\end{equation}

Where $w$ and $b$ are its only learnable parameters and $g_e(\cdot)$ is the decoder edge function applied to every pair of node embeddings. It reflects that edge probabilities will depend on the relative distances among node embeddings. The training loss is defined as the binary cross entropy between the estimated and the ground truth edges $\mathcal{L} = \sum_{ij}\mathrm{BCE}(\hat{A}_{ij}, A_{ij})$.

\begin{figure}[h]
\begin{center}
\centerline{\includegraphics[width=0.7\columnwidth]{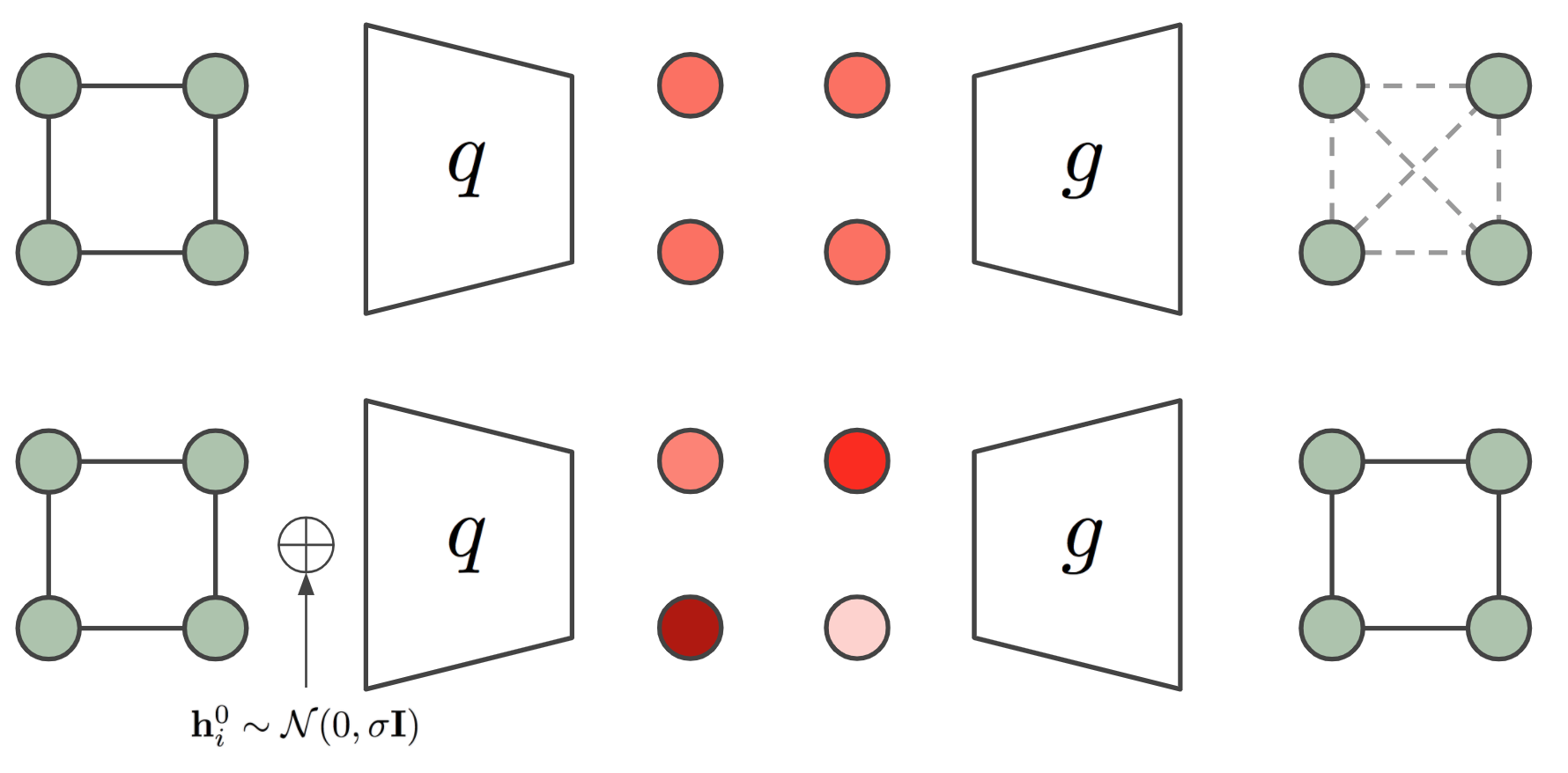}}
\caption{Visual representation of a Graph Autoencoder for a 4 nodes cycle graph. The bottom row illustrates that adding noise at the input graph breaks the symmetry of the embedding allowing the reconstruction of the adjacency matrix.}
\label{fig:symmetric_cycle_graph}
\end{center}
\vskip -0.2in
\end{figure}

\textbf{The symmetry problem:} The above stated autoencoder may seem straightforward to implement at first sight but in some cases there is a strong limitation regarding the symmetry of the graph. Graph Neural Networks are convolutions on the edges and nodes of a graph, i.e. the same function is applied to all edges and to all nodes. In some graphs (e.g. those defined only by its adjacency matrix) we may not have input features in the nodes, and for that reason the difference among nodes relies only on their edges or neighborhood topology. Therefore, if the neighborhood of two nodes is exactly the same, their encoded embeddings will be the same too. A clear example of this is a cycle graph (an example of a 4 nodes cycle graph is provided in Figure \ref{fig:symmetric_cycle_graph}). When running a Graph Neural Network encoder on a node featureless cycle graph, we will obtain the exact same embedding for each of the nodes, which makes it impossible to reconstruct the edges of the original graph from the node embeddings. The cycle graph is a severe example where all nodes have the exact same neighborhood topology but these symmetries can be present in different ways for other graphs with different edge distributions or even when including node features if these are not unique.

  \begin{figure*}[t]
    \centering
    \vspace{-.3cm}
    \begin{minipage}[c]{.59\textwidth}
    \begin{tabular}{lcccccc}
    \toprule
     & \multicolumn{3}{c}{Community Small} &   \multicolumn{3}{c}{Erdos\&Renyi}\\
    Encoder & BCE & \% Error & F1 & BCE & \% Error & F1 \\
    \midrule
    Baseline & - & 31.79 & .0000 & - & 25.13 & 0.000    \\
    GNN & 6.75 & 1.29 & 0.980 & 14.15 & 4.62 & 0.907  \\
    Noise-GNN & 3.32 & 0.44 & 0.993 & 4.56 & 1.25 & 0.975  \\
    Radial Field & 9.22 & 1.19 & 0.981 & 6.78 & 1.63 & 0.968 \\
    EGNN & \textbf{2.14} & \textbf{0.06} & \textbf{0.999} & \textbf{1.65} & \textbf{0.11} & \textbf{0.998}  \\
    \bottomrule
    \end{tabular}
    \end{minipage}\begin{minipage}[c]{.4\textwidth}\vspace{.3cm}
    \centering
    \includegraphics[width=0.99\textwidth]{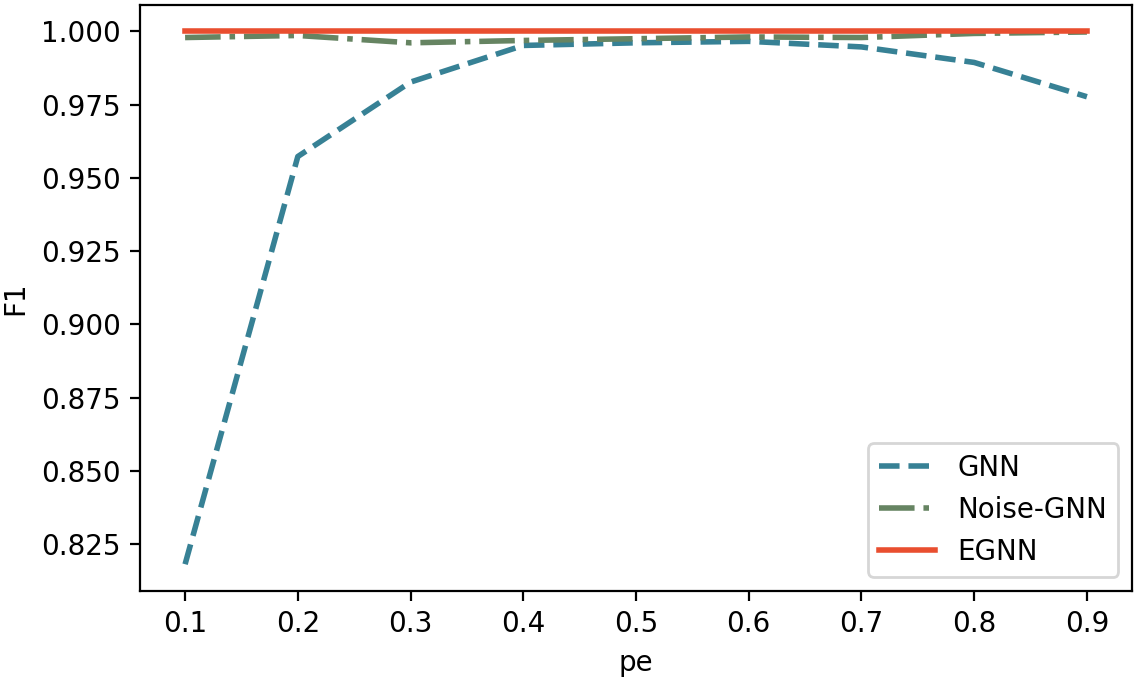}
    \end{minipage}
    \captionlistentry[figure]{}
    \vspace{-.2cm}
    \caption{In the Table at the left we report the Binary Cross Entropy, \% Error and F1 scores for the test partition on the Graph Autoencoding experiment in the Community Small and Erdos\&Renyi datasets. In the Figure at the right, we report the F1 score when overfitting a training partition of 100 samples in the Erdos\&Renyi dataset for different values of sparsity $p_e$. The GNN is not able to successfully auto-encode sparse graphs (small $p_e$ values) for the Erdos\&Renyi dataset even when training and testing on the same small subset.}
    \vspace{-.2cm}
    \label{table:autoencoder_results}
  \end{figure*}

An ad-hoc method to break the symmetry of the graph is introduced by \cite{liu2019graph}. This method introduces noise sampled from a Gaussian distribution into the input node features of the graph $\rmh_i^{0} \sim \mathcal{N}(\mathbf{0}, \sigma\mathbf{I})$. This noise allows different representations for all node embeddings and as a result the graph can be decoded again, but it comes with a drawback, the network has to generalize over the new introduced noise distribution. Our Equivariant Graph Autoencoder will remain translation and rotation equivariant to this sampled noise which we find makes the generalization much easier. Another way of looking at this is considering the sampled noise makes the node representations go from structural to positional  \cite{srinivasan2019equivalence} where $\En$ equivariance may be beneficial. In our case we will simply input this noise as the input coordinates $\rmx^{0} \sim \mathcal{N}(\mathbf{0}, \sigma\mathbf{I}) \in \mathbb{R}^{M \times n}$ of our EGNN which will output an equivariant transformation of them $\rmx^L$, this output will be used as the embedding of the graph (i.e. $\rmz = \rmx^L$) which is the input to the decoder from Equation \ref{eq:graph_decoder}.

\textbf{Dataset:} We generated community-small graphs \cite{you2018graphrnn, liu2019graph} by running the original code from \cite{you2018graphrnn}. These graphs contain $12\leq M \leq20$ nodes. We also generated a second dataset using the Erdos\&Renyi generative model \cite{bollobas2001random} sampling random graphs with an initial number of $7\leq M \leq16$ nodes and edge probability $p_e=0.25$. We sampled $5.000$ graphs for training, $500$ for validation and $500$ for test for both datasets. Each graph is defined as and adjacency matrix $A \in \{0, 1\}^{M \times M}$.

\textbf{Implementation details:} Our Equivariant Graph Auto-Encoder is composed of an EGNN encoder followed by the decoder from Equation \ref{eq:graph_decoder}. The graph edges $A_{ij}$ are input as edge attributes $a_{ij}$ in Equation \ref{eq:method_edge}. The noise used to break the symmetry is input as the coordinates $\rmx^0 \sim \mathcal{N(\mathbf{0}, \sigma\mathbf{I})} \in \mathbb{R}^{M \times n}$ in the first layer and $\rmh^0$ is initialized as ones since we are working with featureless graphs. As mentioned before, the encoder outputs an equivariant transformation on the coordinates which is the graph embedding and input to the decoder $\rmz = \rmx^L \in \mathbb{R}^{M\times n}$. We use $n=8$ dimensions for the embedding space. We compare the EGNN to its GNN cousin, we also compare to Noise-GNN which is an adaptation of our GNN to match the setting from \cite{liu2019graph}, and we also include the Radial Field algorithm as an additional baseline. All four models have 4 layers, 64 features for the hidden layers, the Swish activation function as a non-linearity and they were all trained for 100 epochs using the Adam optimizer and learning rate $10^{-4}$. More details are provided in Appendix \ref{sec:appendix_implementation_graph_autoencoders}. Since the number of nodes is larger than the number of layers, the receptive field of a GNN may not comprise the whole graph which can make the comparison unfair with our EGNN. To avoid this limitation, all models exchange messages among all nodes and the edge information is provided as edge attributes $a_{ij} = A_{ij}$ in all of them.

\textbf{Results:} In the table from Figure \ref{table:autoencoder_results} we report the Binary Cross Entropy loss between the estimated and ground truth edges, the \% Error which is defined as the percentage of wrong predicted edges with respect to the total amount of potential edges, and the F1 score of the edge classification, all numbers refer to the test partition. We also include a "Baseline" that predicts all edges as missing $\hat{A}_{ij}=0$. The standard GNN seems to suffer from the symmetry problem and provides the worst performance. When introducing noise (Noise-GNN), both the loss and the error decrease showing that it is actually useful to add noise to the input nodes. Finally, our EGNN remains $\En$ equivariant to this noise distribution and provides the best reconstruction with a $0.11\%$ error in the Erdos\&Renyi dataset and close to optimal $0.06\%$ in the Community Small dataset. A further analysis of the reconstruction error for different $n$ embedding sizes is reported in Appendix \ref{sec:appendix_graph_autoencoder}.

\begin{table*}[th]
  \centering
  \begin{tabular}{l c c c c c c c c c c c c}
  \toprule
    Task & $\alpha$ & $\Delta \varepsilon$ & $\varepsilon_{\mathrm{HOMO}}$ & $\varepsilon_{\mathrm{LUMO}}$ & $\mu$ & $C_{\nu}$ & $G$ & $H$ & $R^2$ & $U$ & $U_0$ & ZPVE \\
    Units & bohr$^3$ & meV & meV & meV & D & cal/mol K & meV & meV & bohr$^3$ & meV & meV & meV \\
    \midrule
    NMP & .092 & 69 & 43 & 38 & .030 & .040 & 19 & 17 & .180 & 20 & 20 & 1.50 \\
    Schnet & .235 & 63 & 41 & 34 & .033 & .033 & 14 & 14 & .073 & 19 & 14 & 1.70 \\
    Cormorant & .085 & 61 & 34 & 38 & .038 & .026 & 20 & 21 & .961 & 21 & 22 & 2.03  \\
    L1Net & .088 & 68 & 46 & 35 & .043 & .031 & 14 & 14 & .354 & 14 & 13 &  1.56\\
    
    LieConv & .084 & 49 & 30 & 25 & .032 & .038 & 22 & 24 & .800 & 19 & 19 & 2.28 \\
    DimeNet++* & .044 & 33 & 25 & 20 & .030 & .023 & 8 & 7 & .331 & 6 & 6 & 1.21  \\
    TFN & .223 & 58 & 40 & 38 & .064 & .101 & - & - & - & - & - & - \\
    SE(3)-Tr. & .142 & 53 & 35 & 33 & .051 & .054 & - & - & - & - & - & -  \\
    \midrule
    EGNN & .071  & 48 & 29 & 25 & .029 & .031 & 12 & 12 & .106 & 12 & 11 & 1.55  \\
    \bottomrule
  \end{tabular}
  \caption{Mean Absolute Error for the molecular property prediction benchmark in QM9 dataset. *DimeNet++ uses slightly different train/val/test partitions than the other papers listed here.}
  \label{tab:qm9}
\end{table*}

\textbf{Overfitting the training set:} We explained the symmetry problem and we showed the EGNN outperforms other methods in the given datasets. Although we observed that adding noise to the GNN improves the results, it is difficult to exactly measure the impact of the symmetry limitation in these results independent from other factors such as generalization from the training to the test set. In this section we conduct an experiment where we train the different models in a subset of 100 Erdos\&Renyi graphs and embedding size $n=16$ with the aim to overfit the data. We evaluate the methods on the training data. In this experiment the GNN is unable to fit the training data properly while the EGNN can achieve perfect reconstruction and Noise-GNN close to perfect. We sweep over different $p_e$ sparsity values from $0.1$ to $0.9$ since the symmetry limitation is more present in very sparse or very dense graphs. We report the F1 scores of this experiment in the right plot of Figure \ref{table:autoencoder_results}.

In this experiment we showed that $\En$ equivariance improves performance when embedding graphs in a continuous space as a set of nodes in dimension $n$. Even though this is a simple reconstruction task, we think this can be a useful step towards generating graphs or molecules where often graphs (i.e. edges) are decoded as pairwise distances or similarities between nodes e.g. \cite{kipf2016variational, liu2019graph, grover2019graphite}, and these metrics (e.g. eq. \ref{eq:graph_decoder}) are $\En$ invariant. Additionally this experiment also showed that our method can successfully perform in a $\En$ equivariant task for higher dimensional spaces where $n>3$.

\subsection{Molecular data | QM9}
The QM9 dataset \cite{ramakrishnan2014quantum} has become a standard in machine learning as a chemical property prediction task. The QM9 dataset consists of small molecules represented as a set of atoms (up to 29 atoms per molecule), each atom having a 3D position associated and a five dimensional one-hot node embedding that describe the atom type (H, C, N, O, F). The dataset labels are a variety of chemical properties for each of the molecules which are estimated through regression. These properties are invariant to translations, rotations and reflections on the atom positions.
Therefore those models that are E(3) invariant are highly suitable for this task. 

We imported the dataset partitions from \cite{anderson2019cormorant}, 100K molecules for training, 18K for validation and 13K for testing. A variety of 12 chemical properties were estimated per molecule. We optimized and report the Mean Absolute Error between predictions and ground truth.

\textbf{Implementation details:} Our EGNN receives as input the 3D coordinate locations of each atom which are provided as $\rmx_i^0$ in Equation \ref{eq:method_edge} and an embedding of the atom properties which is provided as input node features $\rmh_i^0$.  Since this is an invariant task and also $\rmx^0$ positions are static, there is no need to update the particle's position $\rmx$ by running Equation \ref{eq:method_coords} as we did in previous experiments. Consequently, we tried both manners and we didn't notice any improvement by updating $\rmx$. When not updating the particle's position (i.e. skipping Equation \ref{eq:method_coords}), our model becomes $\mathrm{E}(n)$ invariant, which is analogous to a standard GNN where all relative squared norms between pairs of points $\| \rmx_i - \rmx_j \|^2$ are inputted to the edge operation (eq. \ref{eq:method_edge}). Additionally, since we are not provided with an adjacency matrix and molecules can scale up to 29 nodes, we use the extension of our model from Section \ref{sec:edges_inference} that infers a soft estimation of the edges. Our EGNN network consists of 7 layers, 128 features per hidden layer and the Swish activation function as a non-linearity. A sum-pooling operation preceded and followed by two layers MLPs maps all the node embeddings $\rmh^{L}$ from the output of the EGNN to the estimated property value. Further implementation details are reported in Appendix. \ref{sec:appendix_implementation}. We compare to NMP \cite{gilmer2017neural}, Schnet \cite{schutt2017schnet}, Cormorant \cite{anderson2019cormorant}, L1Net \cite{miller2020relevance}, LieConv \cite{finzi2020generalizing}, DimeNet++ \cite{klicpera2020fast}, TFN \cite{thomas2018tensor} and SE(3)-Tr. \cite{fuchs2020se}.

\textbf{Results} are presented in Table \ref{tab:qm9}. Our method reports very competitive results in all property prediction tasks while remaining relatively simple, i.e. not introducing higher order representations, angles or spherical harmonics. 
Perhaps, surprisingly, we outperform other equivariant networks that consider higher order representations while in this task, we are only using type-0 representations (i.e. relative distances) to define the geometry of the molecules. In Appendix \ref{appendix:invariant_features} we prove that when only positional information is given (i.e. no velocity or higher order type features), then the geometry is completely defined by the norms in-between points up to $\En$-transformations, in other words, if two collections of points separated by $\En$ transformations are considered to be identical, then the relative norms between points is a unique identifier of the collection.

\section{Conclusions}
Equivariant graph neural networks are receiving increasing interest from the natural and medical sciences as they represent a new tool for analyzing molecules and their properties. In this work, we presented a new $\En$ equivariant deep architecture for graphs that is computationally efficient, easy to implement, and significantly improves over the current state-of-the-art on a wide range of tasks. We believe these properties make it ideally suited to make a direct impact on topics such as drug discovery, protein folding and the design of new materials, as well as applications in 3D computer vision.

\section*{Acknowledgements}
We would like to thank Patrick Forr\'e for his support to formalize the invariance features identification proof. 




\bibliography{example_paper}
\bibliographystyle{icml2021}

\onecolumn
\newpage
\appendix 
\section{Equivariance Proof} \label{sec:appendix_equiv_proof}

In this section we prove that our model is translation equivariant on $\rmx$ for any translation vector $g \in \mathbb{R}^n$ and it is rotation and reflection equivariant on $\rmx$ for any orthogonal matrix $Q \in \mathbb{R}^{n \times n}$. More formally, we will prove the model satisfies:

$$Q\rmx^{l+1} + g, \rmh^{l+1}  = \mathrm{EGCL}(Q\rmx^l + g, \rmh^l)$$

We will analyze how a translation and rotation of the input coordinates propagates through our model. We start assuming $\rmh^0$ is invariant to $\En$ transformations on $\rmx$, in other words, we do not encode any information about the absolute position or orientation of $\rmx^0$ into $\rmh^0$. Then, the output $\rmm_{ij}$ of Equation \ref{eq:method_edge} will be invariant too since the distance between two particles is invariant to translations $\|\rmx_{i}^{l} +g - [\rmx_{j}^{l} + g]\|^{2} = \|\rmx_{i}^{l} -\rmx_{j}^{l}\|^{2}$, and it is invariant to rotations and reflections $\|Q\rmx_{i}^{l} - Q\rmx_{j}^{l}\|^{2} = (\rmx_i^l- \rmx_j^l)^\top Q^\top Q (\rmx_{i}^{l} -\rmx_{j}^{l}) = (\rmx_i^l- \rmx_j^l)^\top \mathbf{I} (\rmx_{i}^{l} -\rmx_{j}^{l}) = \|\rmx_{i}^{l} -\rmx_{j}^{l}\|^{2}$ such that the edge operation becomes invariant:

$$\rmm_{i, j} =\phi_{e}\left(\rmh_{i}^{l}, \rmh_{j}^{l},\left\|Q\rmx_{i}^{l} + g- [Q\rmx_{j}^{l} + g]\right\|^{2}, a_{i j}\right) = \phi_{e}\left(\rmh_{i}^{l}, \rmh_{j}^{l},\left\|\rmx_{i}^{l} - \rmx_{j}^{l} \right\|^{2}, a_{i j}\right)$$

The second equation of our model (eq. \ref{eq:method_coords}) that updates the coordinates $\rmx$ is $\En$ equivariant. Following, we prove its equivariance by showing that an $\En$ transformation of the input leads to the same transformation of the output. Notice $\rmm_{ij}$ is already invariant as proven above. We want to show:
$$Q\rmx_{i}^{l+1} + g  = Q\rmx_{i}^{l} + g +C\sum_{j \neq i}\left(Q\rmx_{i}^{l} + g - [Q\rmx_{j}^{l} + g]\right) \phi_{x}\left(\rmm_{i, j}\right)$$

\textit{Derivation.}
\begin{align*}
    Q\rmx_{i}^{l} + g +C\sum_{j \neq i}\left(Q\rmx_{i}^{l} + g - Q\rmx_{j}^{l} - g\right) \phi_{x}\left(\rmm_{i, j}\right)
    &= Q\rmx_{i}^{l} + g +QC\sum_{j \neq i}\left(\rmx_{i}^{l} - \rmx_{j}^{l}\right) \phi_{x}\left(\rmm_{i, j}\right)\\
    &= Q\left(\rmx_{i}^{l} +C\sum_{j \neq i}\left(\rmx_{i}^{l} - \rmx_{j}^{l}\right) \phi_{x}\left(\rmm_{i, j}\right) \right) + g\\
    &= Q\rmx_{i}^{l+1} + g
\end{align*}

Therefore, we have proven that rotating and translating $\rmx^l$ results in the same rotation and translation on $\rmx^{l+1}$ at the output of Equation \ref{eq:method_coords}.

Furthermore equations \ref{eq:method_agg} and \ref{eq:method_node} only depend on $\rmm_{ij}$ and $\rmh^l$ which as saw at the beginning of this proof, are $\En$ invariant, therefore the output of Equation \ref{eq:method_node} $\rmh^{l+1}$ will be invariant too. Thus concluding that a transformation $Q\rmx^l +g$ on $\rmx^l$ will result in the same transformation on $\rmx^{l+1}$ while $\rmh^{l+1}$ will remain invariant to it such that $Q\rmx^{l+1} + g, \rmh^{l+1}  = \mathrm{EGCL}(Q\rmx^l + g, \rmh^l)$ is satisfied.

\section{Re-formulation for velocity type inputs} \label{sec:appendix:method_velocity}

In this section we write down the EGNN transformation layer $\rmh^{l+1}, \rmx^{l+1}, \rmv^{l+1} = \text{EGCL}[\rmh^{l}, \rmx^{l}, \rmv^{\text{init}}, \mathcal{E}]$ that can take in velocity input and output channels. We also prove it remains $\En$ equivariant. %
\begin{align*}
\rmm_{i j} &=\phi_{e}\left(\rmh_{i}^{l}, \rmh_{j}^{l},\left\|\rmx_{i}^{l}-\rmx_{j}^{l}\right\|^{2}, a_{i j}\right) \\ 
 \rmv_{i}^{l+1}&= \phi_{v}\left(\rmh_{i}^l\right)\rmv_{i}^{\text{init}} +C\sum_{j \neq i}\left(\rmx_{i}^{l}-\rmx_{j}^{l}\right) \phi_{x}\left(\rmm_{i j}\right) \\
\rmx_{i}^{l+1} &=\mathbf{x}_{i}^{l}+ \mathbf{v}_i^{l+1} \\
\rmm_{i} &=\sum_{j \neq i} \rmm_{i j} \\
\rmh_{i}^{l+1} &=\phi_{h}\left(\rmh_{i}^l, \rmm_{i}\right)
\end{align*}
\subsection{Equivariance proof for velocity type inputs} \label{sec:appendix_equiv_proof_velocity}
In this subsection we prove that the velocity types input formulation of our model is also $\En$ equivariant on $\rmx$. More formally, for any translation vector $g \in \mathbb{R}^n$ and for any orthogonal matrix $Q \in \mathbb{R}^{n \times n}$, the model should satisfy:
$$\rmh^{l+1}, Q\rmx^{l+1} + g, Q\rmv^{l+1} = \text{EGCL}[\rmh^{l}, Q\rmx^{l} + g, Q\rmv^{\text{init}}, \mathcal{E}]$$

In Appendix \ref{sec:appendix_equiv_proof} we already proved the equivariance of our EGNN (Section \ref{sec:method_main}) when not including vector type inputs. In its velocity type inputs variant we only replaced its coordinate updates (eq. \ref{eq:method_coords}) by Equation \ref{eq:egnn_velocity} that includes velocity. Since this is the only modification we will only prove that Equation \ref{eq:egnn_velocity} re-written below is equivariant.
\begin{align*}
     \rmv_{i}^{l+1}&= \phi_{v}\left(\rmh_{i}^l\right)\rmv_{i}^{\text{init}} +C\sum_{j \neq i}\left(\rmx_{i}^{l}-\rmx_{j}^{l}\right) \phi_{x}\left(\rmm_{i j}\right) \\
\rmx_{i}^{l+1} &=\mathbf{x}_{i}^{l}+ \mathbf{v}_i^{l+1} \\
\end{align*}
First, we prove the first line preserves equivariance, that is we want to show: $$Q\rmv_{i}^{l+1} = \phi_{v}\left(\rmh_{i}^l\right)Q\rmv_{i}^{\text{init}} +C\sum_{j \neq i}\left(Q\rmx_{i}^{l} + g - [Q\rmx_{j}^{l} + g]\right) \phi_{x}\left(\rmm_{i j}\right)$$
\textit{Derivation.}
\begin{align}
    \phi_{v}\left(\rmh_{i}^l\right)Q\rmv_{i}^{\text{init}} +C\sum_{j \neq i}\left(Q\rmx_{i}^{l} + g - [Q\rmx_{j}^{l} + g]\right) \phi_{x}\left(\rmm_{i j}\right)
    &= Q\phi_{v}\left(\rmh_{i}^l\right)\rmv_{i}^{\text{init}} +QC\sum_{j \neq i}\left(\rmx_{i}^{l} - \rmx_{j}^{l}  \right) \phi_{x}\left(\rmm_{i j}\right) \\
        &= Q\left(\phi_{v}\left(\rmh_{i}^l\right)\rmv_{i}^{\text{init}} +C\sum_{j \neq i}\left(\rmx_{i}^{l} - \rmx_{j}^{l}  \right) \phi_{x}\left(\rmm_{i j}\right)\right) \\
        &= Q\rmv_{i}^{l+1}
\end{align}
Finally, it is straightforward to show the second equation is also equivariant, that is we want to show $Q\rmx_{i}^{l+1} + g = Q\mathbf{x}_{i}^{l}+ g + Q\mathbf{v}_i^{l+1}$

\textit{Derivation.}
\begin{align*}
    Q\mathbf{x}_{i}^{l}+ g + Q\mathbf{v}_i^{l+1} &= Q\left (\mathbf{x}_{i}^{l} +\mathbf{v}_i^{l+1} \right ) +g \\
    &= Q\rmx_{i}^{l+1} + g
\end{align*}
Concluding we showed that an $\En$ transformation on the input set of points results in the same transformation on the output set of points such that $\rmh^{l+1}, Q\rmx^{l+1} + g, Q\rmv^{l+1} = \text{EGCL}[\rmh^{l}, Q\rmx^{l} + g, Q\rmv^{\text{init}}, \mathcal{E}]$ is satisfied.

\section{Implementation details} \label{sec:appendix_implementation}
In this Appendix section we describe the implementation details of the experiments.
First, we describe those parts of our model that are the same across all experiments. Our EGNN model from Section \ref{sec:method_main} contains the following three main learnable functions.

\begin{itemize}
    \item \textbf{The edge function} $\phi_e$ (eq. \ref{eq:method_edge}) is a two layers MLP with two Swish non-linearities: Input $\xrightarrow{}$ \{LinearLayer() $\xrightarrow{}$ Swish() $\xrightarrow{}$ LinearLayer() $\xrightarrow{}$ Swish() \} $\xrightarrow{}$ Output. 
    \item \textbf{The coordinate function} $\phi_x$ (eq. \ref{eq:method_coords}) consists of a two layers MLP with one non-linearity:  $\rmm_{ij}$ $\xrightarrow{}$ \{LinearLayer() $\xrightarrow{}$ Swish() $\xrightarrow{}$ LinearLayer() \} $\xrightarrow{}$ Output
    \item \textbf{The node function} $\phi_h$ (eq. \ref{eq:method_node}) consists of a two layers MLP with one non-linearity and a residual connection:
    
    [$\rmh_i^{l}$, $\rmm_i$] $\xrightarrow{}$ \{LinearLayer() $\xrightarrow{}$ Swish() $\xrightarrow{}$ LinearLayer() $\xrightarrow{}$ Addition($\rmh^l_i$) \} $\xrightarrow{}$ $\mathbf{h}^{l+1}_i$ 
\end{itemize}

These functions are used in our EGNN across all experiments. Notice the GNN (eq. \ref{eq:gnn}) also contains and edge operation and a node operation $\phi_e$ and $\phi_h$ respectively. We use the same functions described above for both the GNN and the EGNN such that comparisons are as fair as possible.

\subsection{Implementation details for Dynamical Systems}
\label{sec:appendix_implementation_dynamical}
\textbf{Dataset}

In the dynamical systems experiment we used a modification of the Charged Particle's N-body (N=5) system  from \cite{kipf2018neural}. Similarly to \cite{fuchs2020se}, we extended it from 2 to 3 dimensions customizing the original code from  \hyperlink{https://github.com/ethanfetaya/NRI}{(https://github.com/ethanfetaya/NRI)} and we removed the virtual boxes that bound the particle's positions. The sampled dataset consists of 3.000 training trajectories, 2.000 for validation and 2.000 for testing. Each trajectory has a duration of 1.000 timesteps. To move away from the transient phase, we actually generated trajectories of 5.000 time steps and sliced them from timestep $3.000$ to timestep $4.000$ (1.000 time steps into the future) such that the initial conditions are more realistic than the Gaussian Noise initialization from which they are initialized. 

In our second experiment, we sweep from 100 to 50.000 training samples, for this we just created a new training partition following the same procedure as before but now generating 50.000 trajectories instead. The validation and test partition remain the same from last experiment.

\textbf{Models}

All models are composed of 4 layers, the details for each model are the following.
\begin{itemize}
    \item \textbf{EGNN}: For the EGNN we use its variation that considers vector type inputs from Section \ref{sec:method_velocity}. This variation adds the function $\phi_v$ to the model which is composed of two linear layers with one non-linearity: Input $\xrightarrow{}$ \{LinearLayer() $\xrightarrow{}$ Swish() $\xrightarrow{}$ LinearLayer() \} $\xrightarrow{}$ Output. Functions $\phi_e$, $\phi_x$ and $\phi_h$ that define our EGNN are the same than for all experiments and are described at the beginning of this Appendix \ref{sec:appendix_implementation}.
    \item \textbf{GNN}: The GNN is also composed of 4 layers, its learnable functions edge operation $\phi_e$ and node operation $\phi_h$ from Equation \ref{eq:gnn} are exactly the same as $\phi_e$ and $\phi_h$ from our EGNN introduced in Appendix \ref{sec:appendix_implementation}. We chose the same functions for both models to ensure a fair comparison. In the GNN case, the initial position $\rmp^0$ and velocity $\rmv^0$ from the particles is passed through a linear layer and inputted into the GNN first layer $\rmh^0$. The particle's charges are inputted as edge attributes $a_{ij} = c_i c_j$. The output of the GNN $\rmh^L$ is passed through a two layers MLP that maps it to the estimated position.
    
    \item \textbf{Radial Field}: The Radial Field algorithm is described in the Related Work \ref{sec:related_work}, its only parameters are contained in its edge operation $\phi_{\mathrm{rf}}()$ which in our case is a two layers MLP with two non linearities Input $\xrightarrow{}$ \{LinearLayer() $\xrightarrow{}$ Swish() $\xrightarrow{}$ LinearLayer() $\xrightarrow{}$ Tanh \} $\xrightarrow{}$ Output. Notice we introduced a Tanh at the end of the MLP which fixes some instability issues that were causing this model to diverge in the dynamical system experiment. We also augmented the Radial Field algorithm with the vector type inputs modifications introduced in Section \ref{sec:method_velocity}. In addition to the norms between pairs of points, $\phi_{\mathrm{rf}}()$ also takes as input the particle charges $c_i c_j$.
    
    \item \textbf{Tensor Field Network:} We used the Pytorch implementation from  \hyperlink{https://github.com/FabianFuchsML/se3-transformer-public}{https://github.com/FabianFuchsML/se3-transformer-public}. We swept over different hyper paramters, degree $\in$ \{2, 3, 4\}, number of features $\in$ \{12, 24, 32, 64, 128\}. We got the best performance in our dataset for degree 2 and number of features 32. We used the Relu activation layer instead of the Swish for this model since it provided better performance.
    
    \item \textbf{SE(3) Transformers:} For the SE(3)-Transformer we used code from \hyperlink{https://github.com/FabianFuchsML/se3-transformer-public}{https://github.com/FabianFuchsML/se3-transformer-public}. Notice this implementation has only been validated in the QM9 dataset but it is the only available implementation of this model. We swept over different hyperparamters degree $\in$ \{1, 2, 3, 4\}, number of features $\in$ 16, 32, 64 and divergence $\in$ \{1, 2\}, along with the learning rate. We obtained the best performance for degree 3, number of features 64 and divergence 1. As in Tensor Field Networks we obtained better results by using the Relu activation layer instead of the Swish.
\end{itemize}

\textbf{Other implementation details}

In Table \ref{table:n_body} all models were trained for 10.000 epochs, batch size 100, Adam optimizer, the learning rate was fixed and independently chosen for each model. All models are 4 layers deep and the number of training samples was set to 3.000.

\subsection{Implementation details for Graph Autoneoders}
\label{sec:appendix_implementation_graph_autoencoders}

\textbf{Dataset}

In this experiment we worked with Community Small \cite{you2018graphrnn} and Erdos\&Renyi \cite{bollobas2001random} generated datasets.

\begin{itemize}
    \item Community Small: We used the original code from \cite{you2018graphrnn} (\hyperlink{https://github.com/JiaxuanYou/graph-generation}{https://github.com/JiaxuanYou/graph-generation}) to generate a Community Small dataset. We sampled 5.000 training graphs, 500 for validation and 500 for testing.
    \item Erdos\&Renyi is one of the most famous graph generative algorithms. We used the "gnp\_random\_graph($M$, $p$)" function from (\hyperlink{https://networkx.org/}{https://networkx.org/}) that generates random graphs when povided with the number of nodes $M$ and the edge probability $p$ following the Erdos\&Renyi model. Again we generated 5.000 graphs for training, 500 for validation and 500 for testing. We set the edge probability (or sparsity value) to $p=0.25$ and the number of nodes $M$ ranging from 7 to 16 deterministically uniformly  distributed. Notice that edges are generated stochastically with probability $p$, therefore, there is a chance that some nodes are left disconnected from the graph, "gnp\_random\_graph($M$, $p$)" function discards these disconnected nodes such that even if we generate graphs setting parameters to $7 \leq M \leq 16$ and $p=0.25$ the generated graphs may have less number of nodes. \end{itemize}
    
Finally, in the graph autoencoding experiment we also overfitted in a small partition of 100 samples (Figure \ref{table:autoencoder_results}) for the Erdos\&Renyi graphs described above. We reported results for different $p$ values ranging from $0.1$ to $0.9$. For each $p$ value we generated a partition of 100 graphs with initial number of nodes between $7 \leq M \leq 16$ using the Erdos\&Renyi generative model.

\textbf{Models}

All models consist of 4 layers, 64 features for the hidden layers and the Swish activation function as a non linearity. The EGNN is defined as explained in Section \ref{sec:method_main} without any additional modules (i.e. no velocity type features or inferring edges). The functions $\phi_e$, $\phi_x$ and $\phi_h$ are defined at the beginning of this Appendix \ref{sec:appendix_implementation}. The GNN (eq. \ref{eq:gnn}) mimics the EGNN in terms that it uses the same $\phi_h$ and $\phi_e$ than the EGNN for its edge and node updates. The Noise-GNN is exactly the same as the GNN but inputting noise into the $\rmh_0$ features. Finally the Radial Field was defined in the Related Related work Section \ref{sec:related_work} which edge's operation $\phi_{\mathrm{rf}}$ consists of a two layers MLP: Input $\xrightarrow{}$ \{ Linear() $\xrightarrow{}$ Swish() $\xrightarrow{}$ Linear() \} $\xrightarrow{}$ Output.

\textbf{Other implementation details}

All experiments have been trained with learning rate $10^{-4}$, batch size 1, Adam optimizer,  weight decay $10^{-16}$, 100 training epochs for the 5.000 samples sized datasets performing early stopping for the minimum Binary Cross Entropy loss in the validation partition. The overfitting experiments were trained for 10.000 epochs on the 100 samples subsets.

\subsection{Implementation details for QM9}
\label{sec:appendix_implementation_qm9_dataset}
For QM9 \cite{ramakrishnan2014quantum} we used the dataset partitions from \cite{anderson2019cormorant}. We imported the dataloader from his code repository (\hyperlink{https://github.com/risilab/cormorant}{https://github.com/risilab/cormorant}) which includes his data-preprocessing. Additionally all properties have been normalized by substracting the mean and dividing by the Mean Absolute Deviation.

Our EGNN consists of 7 layers. Functions $\phi_e$ and $\phi_h$ are defined at the beginning of this Appendix \ref{sec:appendix_implementation}. Additionally, we use the module $\phi_{inf}$ presented in Section \ref{sec:edges_inference} that infers the edges . This function $\phi_{inf}$ is defined as a linear layer followed by a sigmoid: Input $\xrightarrow{}$ \{Linear() $\xrightarrow{}$ sigmoid()\} $\xrightarrow{}$ Output. Finally, the output of our EGNN $\rmh^L$ is forwarded through a two layers MLP that acts node-wise, a sum pooling operation and another two layers MLP that maps the averaged embedding to the predicted property value, more formally: $\rmh^L$ $\xrightarrow{}$ \{Linear() $\xrightarrow{}$ Swish() $\xrightarrow{}$ Linear() $\xrightarrow{}$ Sum-Pooling() $\xrightarrow{}$ Linear() $\xrightarrow{}$ Swish() $\xrightarrow{}$ Linear\} $\xrightarrow{}$ Property. The number of hidden features for all model hidden layers is 128.

We trained each property individually for a total of 1.000 epochs, we used Adam optimizer, batch size 96, weight decay $10^{-16}$, and cosine decay for the learning rate starting at at a lr=$5\cdot10^{-4}$ except for the Homo, Lumo and Gap properties where its initial value was set to $10^{-3}$.

\section{Further experiments}
\subsection{Graph Autoencoder}  \label{sec:appendix_graph_autoencoder}
In this section we present an extension of the Graph Autoencoder experiment \ref{sec:expriment_autoencoder}. In Table \ref{tab:appendix_graph_autoencoder} we report the approximation error of the reconstructed graphs as the embedding dimensionality $n$ is reduced $n \in \{4, 6, 8 \}$ in the Community Small and Erdos\&Renyi datasets for the GNN, Noise-GNN and EGNN models. For small embedding sizes ($n=4$) all methods perform poorly, but as the embedding size grows our EGNN significantly outperforms the others.

\begin{table}[]
\renewcommand{\arraystretch}{0.5}
\begin{center} \small
\begin{tabular}{l | cc | cc | cc | cc | cc | cc  } 
    \toprule
     & \multicolumn{6}{|c}{Community Small} &   \multicolumn{6}{|c}{Erdos\&Renyi}\\
     \midrule
     & \multicolumn{2}{|c|}{n=4} &  \multicolumn{2}{|c|}{n=6} & \multicolumn{2}{|c|}{n=8} &
      \multicolumn{2}{|c|}{n=4} &  \multicolumn{2}{|c|}{n=6} & \multicolumn{2}{|c}{n=8} \\
     \midrule
      & \% Err. & F1  & \% Err. & F1 & \% Err. & F1 & \% Err. & F1 & \% Err. & F1 & \% Err. & F1 \\
    \midrule
    GNN & \textbf{1.45} & \textbf{0.977} & 1.29 & 0.9800 & 1.29 & 0.980 & 7.92 & 0.844 & 5.22 & 0.894 & 4.62 & 0.907   \\
    Noise-GNN & 1.94 & 0.970 & 0.44 & 0.9931 & 0.44 & 0.993 & 3.80 & 0.925 & 2.66 & 0.947 & 1.25 & 0.975   \\
    EGNN & 2.19 & 0.966 & \textbf{0.42} & \textbf{0.9934} & \textbf{0.06} & \textbf{0.999} & \textbf{3.09} & \textbf{0.939} & \textbf{0.58} & \textbf{0.988} & \textbf{0.11} & \textbf{0.998}   \\
    \bottomrule
\end{tabular}
\caption{Analysis of the \% of wrong edges and F1 score for different $n$ embedding sizes \{2, 4, 8 \} for the GNN, Noise-GNN and EGNN in Community Small and Erdos\&Renyi datasets.}
\label{tab:appendix_graph_autoencoder} 
\end{center}
\end{table}

\newpage
\section{Sometimes invariant features are all you need.} \label{appendix:invariant_features} 
Perhaps surprisingly we find our EGNNs outperform other equivariant networks that consider higher-order representations. In this section we prove that when only positional information is given (i.e. no velocity-type features) then the geometry is completely defined by the invariant distance norms in-between points, without loss of relevant information. As a consequence, it is not necessary to consider higher-order representation types of the relative distances, not even the relative differences as vectors. To be precise, note that these invariant features still need to be \textit{permutation} equivariant, they are only $\En$ invariant.

To be specific, we want to show that for a collection of points $\{\rmx_i\}_{i=1}^M$ the norm of in-between distances $\ell_2(\rmx_i, \rmx_j)$ are a \textit{unique} identifier of the geometry, where collections separated by an $\En$ transformations are considered to be identical. We want to show \textit{invariance} of the norms under $\En$ transformations and \textit{uniqueness}: two point collections are identical (up to $\En$ transform) when they have the same distance norms.

\textbf{\textit{Invariance}}. Let $\{\rmx_i\}$ be a collection of $M$ points where $\rmx_i \in \mathbb{R}^n$ and the $\ell_2$ distances are $\ell_2(\rmx_i, \rmx_j)$. We want to show that all $\ell_2(\rmx_i, \rmx_j)$ are unaffected by $\En$ transformations.

\textit{Proof}. Consider an arbitrary $\En$ transformation $\mathbb{R}^n \to \mathbb{R}^n : \rmx \mapsto Q\rmx + t$ where $Q$ is orthogonal and $t \in \mathbb{R}^n$ is a translation. Then for all $i, j$: \vspace{-.2cm}
\small\begin{align*}
    \ell_2(Q\rmx_i + t, Q\rmx_j + t) &= \sqrt{} (Q\rmx_i + t - [Q\rmx_j + t])^T(Q\rmx_i + t - [Q\rmx_j + t]) = \sqrt{} (Q\rmx_i - Q\rmx_j)^T(Q\rmx_i - Q\rmx_j) \\
    &= \sqrt{} (\rmx_i - \rmx_j)^T Q^T Q (\rmx_i - \rmx_j) = \sqrt{} (\rmx_i - \rmx_j)^T (\rmx_i - \rmx_j) = \ell_2(\rmx_i, \rmx_j)
\end{align*}
\normalsize
This proves that the $\ell_2$ distances are invariant under $\En$ transforms.

\textbf{\textit{Uniqueness}}. Let $\{\rmx_i\}$ and $\{\rmy_i\}$ be two collection of $M$ points each where all in-between distance norms are identical, meaning $\ell_2(\rmx_i, \rmx_j) = \ell_2(\rmy_i, \rmy_j)$. We want to show that $\rmx_i = Q \rmy_i + t$ for some orthogonal $Q$ and translation $t$, for all $i$.

\textit{Proof.} Subtract $\rmx_0$ from all $\{\rmx_i\}$ and $\rmy_0$ from all $\{\rmy_i\}$, so $\tilde{\rmx}_i = \rmx_i - \rmx_0$ and $\tilde{\rmy}_i = \rmy_i - \rmy_0$. As proven above, since translation is an $\En$ transformation the distance norms are unaffected and:
\small
\begin{equation*}
    \ell_2(\tilde{\rmx}_i, \tilde{\rmx}_j) = \ell_2(\rmx_i, \rmx_j) = \ell_2(\rmy_i, \rmy_j) = \ell_2(\tilde{\rmy}_i, \tilde{\rmy}_j).
\end{equation*}\normalsize
So without loss of generality, we may assume that $\rmx_0 = \rmy_0 = \mathbf{0}$. As a direct consequence $||\rmx_i||_2 = ||\rmy_i||_2$. Now writing out the square:
\small
\begin{equation*}
    \rmx_i^T\rmx_i - 2 \rmx_i^T\rmx_j + \rmx_j^T\rmx_j = ||\rmx_i - \rmx_j||_2^2 = ||\rmy_i - \rmy_j||_2^2 = \rmy_i^T\rmy_i - 2 \rmy_i^T\rmy_j + \rmy_j^T\rmy_j
\end{equation*} \normalsize
And since $||\rmx_i||_2 = ||\rmy_i||_2$, it follows that $\rmx_i^T\rmx_j = \rmy_i^T\rmy_j$ or equivalently written as dot product $\langle \rmx_i, \rmx_j\rangle = \langle \rmy_i, \rmy_j \rangle$. Notice that this already shows that angles between pairs of points are the same.

At this moment, it might already be intu\"{i}tive that the collections of points are indeed identical. To finalize the proof formally we will construct a linear map $A$ for which we will show that (1) it maps every $\rmx_i$ to $\rmy_i$ and (2) that it is orthogonal. First note that from the angle equality it follows immediately that for every linear combination:
\small 
\begin{equation*}
    || \sum_i c_i \rmx_i ||_2 = || \sum_i c_i \rmy_i ||_2 \quad (*).\vspace{-.3cm}
\end{equation*}\normalsize
Let $V_x$ be the linear span of $\{\rmx_i\}$ (so $V_x$ is the linear subspace of all linear combinations of $\{\rmx_i\}$). Let $\{\rmx_{i_j}\}_{j=1}^d$ be a basis of $V_x$, where $d \leq n$. Recall that one can define a linear map by choosing a basis, and then define for each basis vector where it maps to. Define a linear map $A$ from $V_x$ to $V_y$ by the transformation from the basis $\rmx_{i_j}$ to $\rmy_{i_j}$ for $j=1,...,d$.
Now pick any point $\rmx_i$ and write it in its basis $\rmx_i = \sum_j c_j \rmx_{i_j} \in V_x$. We want to show $A\rmx_i = \rmy_i$ or alternatively $||\rmy_i - A\rmx_i||_2 = 0$. Note that $A\rmx_i = A \sum_j c_j \rmx_{i_j} =  \sum_j c_j A \rmx_{i_j} = \sum_j c_j \rmy_{i_j}$. Then:
\small
\begin{align*}
    ||\rmy_i - \sum_j c_j \rmy_{i_j}||_2^2 &= \langle \rmy_i, \rmy_i \rangle - 2 \langle \rmy_i, \sum_j c_i \rmy_{i_j} \rangle + \langle \sum_j c_i \rmy_{i_j}, \sum_j c_i \rmy_{i_j} \rangle \\
    &\stackrel{(*)}{=} \langle \rmx_i, \rmx_i \rangle - 2 \langle \rmx_i, \sum_j c_i \rmx_{i_j} \rangle + \langle \sum_j c_i \rmx_{i_j}, \sum_j c_i \rmx_{i_j} \rangle = \langle \rmx_i, \rmx_i \rangle - 2 \langle \rmx_i, \rmx_i \rangle + \langle \rmx_i, \rmx_i \rangle = 0.
\end{align*}\normalsize
Thus showing that $A\rmx_i = \rmy_i$ for all $i = 1, \ldots, M$, proving (1). Finally we want to show that $A$ is orthogonal, when restricted to $V_x$. This follows since:
\begin{equation*}
    \langle A \rmx_{i_j}, A \rmx_{i_j} \rangle = \langle \rmy_{i_j}, \rmy_{i_j} \rangle = \langle \rmx_{i_j}, \rmx_{i_j} \rangle 
\end{equation*}
for the basis elements $\rmx_{i_1},...,\rmx_{i_d}$. This implies that $A$ is orthogonal (at least when restricted to $V_x$). Finally $A$ can be extended via an orthogonal complement of $V_x$ to the whole space. This concludes the proof for (2) and shows that $A$ is indeed orthogonal.

\end{document}